%% file: iros_2017_crf3d.tex
\documentclass[letterpaper, 10 pt, conference]{ieeeconf}  

\IEEEoverridecommandlockouts                              

\usepackage{graphics} 
\usepackage{amsmath} 
\usepackage{cite}
\usepackage{subfigure}
\usepackage{tabulary}
\usepackage{multirow}
\usepackage{todonotes}
\graphicspath{{figs/}}
\usepackage[colorlinks,linkcolor=blue,anchorcolor=blue,citecolor=red]{hyperref} 
\usepackage{wrapfig,lipsum,booktabs}
\usepackage{xcolor}
\usepackage{color}
\usepackage{cite}
\usepackage{algorithmic} 
\usepackage[ruled,vlined,commentsnumbered]{algorithm2e}

\usepackage[flushleft]{threeparttable}
\usepackage{gensymb}  
\usepackage{amsmath}
\usepackage{amssymb}

\usepackage{flushend}

\graphicspath{{figs/}}

\usepackage{siunitx}
\usepackage{url}   

\title{\LARGE \bf
Semantic 3D Occupancy Mapping through Efficient High Order CRFs
}

\author{Shichao Yang$^\star$, Yulan Huang$^*$ and Sebastian Scherer$^\star$ 
\thanks{$^\star$ The Robotics Institute, Carnegie Mellon University, 5000 Forbes Ave, Pittsburgh, PA 15213, USA.
		{\tt\small \{shichaoy, basti\}@andrew.cmu.edu}		}
\thanks{$*$ Language Technology Institute, Carnegie Mellon University, {\tt\small yulanh@andrew.cmu.edu} }		
}

\begin{document}

\maketitle
\thispagestyle{empty}
\pagestyle{empty}

\begin{abstract}
    \input{abstract}
\end{abstract}

\section{Introduction}
\label{sec:kitti crf intro}
\input{intro}

\section{Related Work}
\label{sec:kitti crf review}
\input{related}

\section{Geometric Mapping}
\label{sec:kitti crf geometry map}
\input{3d_reconstruction}

\section{Hierarchical Semantic Mapping}
\label{sec:kitti crf semantic map}
\input{crf_model}

\section{Experiments}
\label{sec:kitti crf experiments}
\input{result}

\section{Conclusion and Future Work}
\label{sec:kitti crf conclusion}
\input{conclusion}


\section*{ACKNOWLEDGMENTS}

The authors would like to thank Daniel Maturana for the scrolling grid mapping implementation \cite{scrollgrid}.

\bibliographystyle{unsrt}    
\bibliography{ref}

\addtolength{\textheight}{-12cm}   

\end{document}

%% file: abstract.tex
Semantic 3D mapping can be used for many applications such as robot navigation and virtual interaction. In recent years, there has been great progress in semantic segmentation and geometric 3D mapping. However, it is still challenging to combine these two tasks for accurate and large-scale semantic mapping from images. In the paper, we propose an incremental and (near) real-time semantic mapping system.
A 3D scrolling occupancy grid map is built to represent the world, which is memory and computationally efficient and bounded for large scale environments. We utilize the CNN segmentation as prior prediction and further optimize 3D grid labels through a novel CRF model. Superpixels are utilized to enforce smoothness and form robust $P^N$ high order potential. An efficient mean field inference is developed for the graph optimization. We evaluate our system on the KITTI dataset and improve the segmentation accuracy by 10\% over existing systems.

%% file: intro.tex
3D semantic mapping is important for many robot applications such as autonomous navigation and robot interaction. Robots not only need to build 3D geometric maps of the environments to avoid obstacles, but also need to recognize objects and scenes for high-level tasks. For example, autonomous vehicles need to locate and also classify vehicles and pedestrians in 3D space to keep safe. However, there are also many challenges with this task. Instead of offline batch optimization, it should be processed incrementally in real time rates and computation time should be independent of the size of environments.


The problem is composed of two parts: geometric reconstruction, and semantic segmentation. For 3D reconstruction, there has been a large amount of research on visual simultaneous localization and mapping (SLAM). The map can be composed of different geometric elements such as points \cite{mur2015orb}, planes\cite{syang2016popslam}, and grid voxels \cite{newcombe2011kinectfusion} etc. For semantic segmentation, current research usually focuses on image or video segmentation for 2D pixel labeling. With the popularity of convolutional neural networks (CNN), the performance of 2D segmentation has greatly improved. However, 2D semantic reasoning is still not accurate in the case of occlusion and shadowing.

Recently, there has also been some work on semantic 3D reconstruction \cite{kundu2014joint} \cite{vineet2015incremental}. However, most of the existing approaches suffer from a variety of limitations. For example, they cannot run incrementally in real time and cannot adapt to large scale scenarios. Some system can achieve real-time rates with GPU acceleration \cite{vineet2015incremental}. Different sensors can be used to accomplish the task such as RGBD cameras \cite{hermans2014dense}, however, they can only work indoors in small workspaces, therefore we choose stereo cameras due to their wide applicability to both indoor and outdoor environments.


In this work, we utilize CNN model to compute pixel label distributions from 2D image and transfer it to 3D grid space. We then propose a Conditional Random Field (CRF) model with higher order cliques to enforce semantic consistency among grids. The clique is generated through superpixels. We develop an efficient filter based mean field approximation inference for this hierarchical CRF. To be applicable to large-scale environments and to achieve real-time computation, a scrolling occupancy grid is built to represent the world which is memory and computationally bounded. In all, our main contributions are:
\begin{itemize}
\item Propose a (near) real-time incremental semantic 3D mapping system for large-scale environments using a scrolling occupancy grid
\item Improve the segmentation accuracy by 10\% over the state-of-the art systems on the KITTI dataset
\item Develop a filter based mean filed inference for high order CRFs with robust $P^n$ potts model by transforming it into a hierarchical pairwise model
\end{itemize}

\begin{figure}[t]
  \centering     
  \label{fig:robust pn}
  \includegraphics[width=3.0in,height=2.0in]{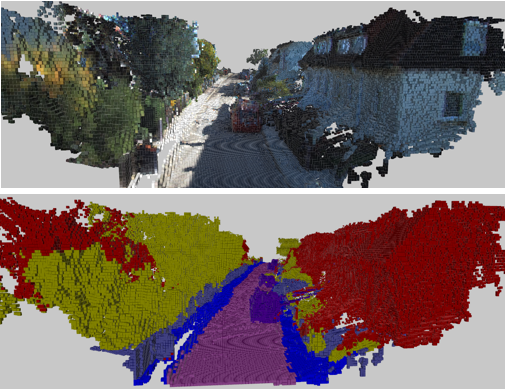}
  \includegraphics[width=3.2in,height=0.12in]{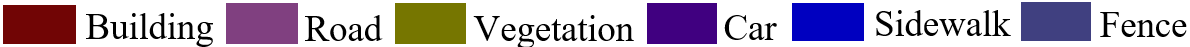}  
   \caption{(top) Geometric and (bottom) semantic 3D reconstruction. Our system can incrementally create semantic map of large scale environments using scrolling occupancy grids in (near) real time. We utilize the latest CNN and build a novel hierarchical CRF model to optimize the label.}
\end{figure}

The paper is organized as follows. Section \ref{sec:kitti crf review} presents some related work. The system contains two main parts: geometric mapping in Section \ref{sec:kitti crf geometry map} and semantic mapping in Section \ref{sec:kitti crf semantic map}. Results and conclusions are presented in Section \ref{sec:kitti crf experiments} and Section \ref{sec:kitti crf conclusion}.

%% file: related.tex
In this section, we first give an overview of the geometric 3D reconstruction and semantic reconstruction, and then introduce some work of the CRF optimization.

\subsection{3D reconstruction}
There are many different algorithms for geometric 3D mapping in recent years such as ORB SLAM \cite{mur2015orb}, and DSO \cite{engel2016direct}, which can achieve impressive results in normal environments. Their maps are usually composed of points, lines or planes. Since points are continuous in space, it is time-consuming and even intractable to perform inference each point's label. A typical approach is to divide the space into discrete grids such as an occupancy grid \cite{fang2016robust} \cite{scherer2012river} and octomap \cite{hornung2013octomap}, which have already been used in many dense and semantic mapping algorithms \cite{vineet2015incremental}.

\subsection{Semantic reconstruction}
A summary of the relevant semantic 3D mapping work is provided in Table \ref{table:kitti crf comparison}. A straightforward solution is to directly transfer the 2D image label to 3D by back-projection \cite{sengupta2013urban} without further 3D optimization. Hermans \textit{et al.} \cite{hermans2014dense} propose to optimize 3D label through a dense pairwise CRF. Vineet \textit{et at.} \cite{vineet2015incremental} apply the same CRF model to large scale stereo 3D reconstruction and achieve real-time rates by GPU. Recently, more complicated high order CRF models are also used for 3D reasoning. Kundu \textit{et al.} \cite{kundu2014joint} model voxel's occupancy and label in one unified CRF with high order ray factors. The super-voxels in octomap \cite{sengupta2015semantic} or 2D superpixel \cite{zhao2016building} can also be used to form high order potential.

\begin{table}
\vspace{1.5 mm}
\caption{Comparison with other system}
\begin{center}
\begin{tabular}{c | c c c c c}
\hline
\multirow{2}{*}{Method}		& Images   	&\multirow{2}{*}{Incremental}     &High-order     & \multirow{2}{*}{Real-time}  \\ 
							& only       &                              & CRF      & \\ \hline
Sengupta  \cite{sengupta2013urban}      &\checkmark &             &	        		&  \\
Hermans \cite{hermans2014dense}		   &           &\checkmark   &       	    &\checkmark    \\
Kundu \cite{kundu2014joint}             &\checkmark &\checkmark   &\checkmark 	& \\
Sengupta  \cite{sengupta2015semantic}   &\checkmark &\checkmark   &\checkmark 	&\\
Vinnet  \cite{vineet2015incremental}    &\checkmark &\checkmark   &    	    		&\checkmark\\
Zhao  \cite{zhao2016building}           &           &    	         &\checkmark 	&--   \\
Li \cite{li2016semi}           		   &\checkmark &\checkmark   & 				&\checkmark   \\
Ours									   &\checkmark & \checkmark  &\checkmark  	& \checkmark\\
\hline
\end{tabular}
\end{center}
\label{table:kitti crf comparison}
\end{table}

\subsection{CRF Inference}
Recently, \emph{dense CRFs} \cite{krahenbuhl2012efficient} have become a popular tool for semantic segmentation and have been applied to many systems \cite{vineet2015incremental} \cite{zhao2016building}. Dense CRFs can model long range relationships compared to a basic neighbour connected CRF model. More complicated CRF with \emph{high order potentials} can be used to encourage label consistency within one region. Since exact inferences for CRFs is generally intractable, many \emph{approximation algorithms} have been developed, such as variants of belief propagation or mean field approximation for dense CRFs \cite{krahenbuhl2012efficient}. Vineet \textit{et al.} \cite{vineet2014filter} extend this inference to CRFs with a $P^n$ Potts model, applied to 3D semantic mapping system\cite{zhao2016building}.

%% file: 3d_reconstruction.tex
\begin{figure*}
\vspace{1.5 mm}
    \centering
    \includegraphics[scale=0.65]{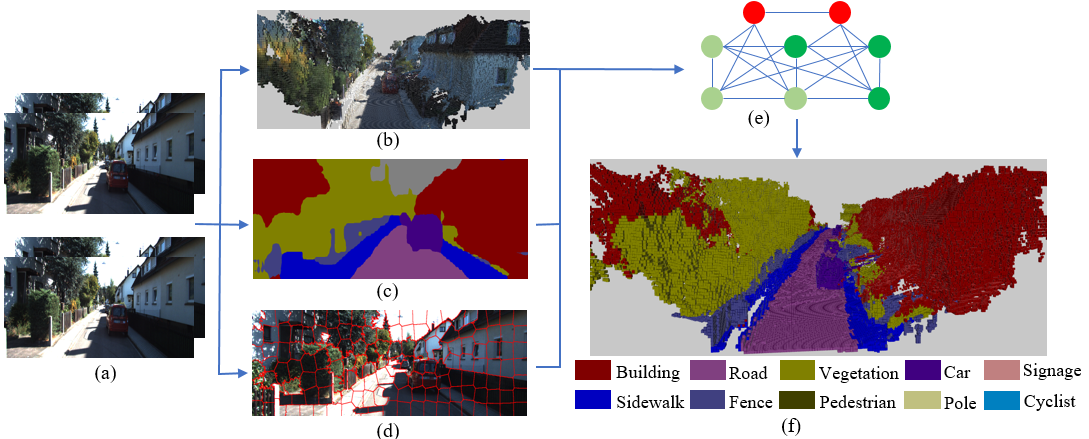}
    \caption{Overview of our system. (a) Input stereo image pairs. (b) Geometric 3D reconstruction using ORB SLAM \cite{mur2015orb}   and occupancy mapping \cite{scherer2012river}\cite{fang2016robust}. (c) 2D semantic segmentation using CNN \cite{yu2015multi}, treated as the CRF unary potential. (d) Superpixel clique generation using SLIC \cite{achanta2012slic} to enforce label consistency within a region, to calculate CRF high order potential. (e) Proposed hierarchical CRF model. We develop efficient mean field inference for hierarchical CRF in 3D grid space. (f) Final semantic mapping after 3D CRF optimization.
} 
    \label{fig:kitti crf whole system}
\end{figure*}

The system diagram is shown in Fig. \ref{fig:kitti crf whole system}. The input to our system is a series of stereo pair images and the output is an incrementally constructed 3D semantic grid map. We break the problem into two steps: geometric 3D mapping (this section) and semantic 3D labelling (next section).


\subsection{3D mapping}
\label{sec: kitti crf 3d mapping}
We first build a 3D geometric map from stereo image pairs which contains three steps: stereo depth estimation, camera pose estimation, and 3D grid mapping.

In order to achieve a dense 3D mapping, we need to accurately estimate stereo disparity with high density. We adopt ELAS \cite{geiger2010efficient} which forms a triangulation on a set of support points to robustly estimate disparity in low-texture areas. We then utilize stereo ORB SLAM \cite{mur2015orb} to estimate 6DoF camera pose. It detects the ORB feature points and then minimizes the reprojection error across different views.

Since ORB SLAM only generates a sparse map of feature points which cannot be used for dense mapping, we project each frame's estimated dense disparity to 3D space. In order to fuse the observations across different views, we change the point cloud map into 3D occupancy grids \cite{fang2016robust}. Each grid stores the probability of being occupied and is updated incrementally through ray tracing based on stereo depth measurement. To keep memory and computation efficient, the occupancy map keeps a fixed dimension and moves along with the camera. If the occupancy value exceeds a threshold, the grid will be considered as occupied and considered for the latter CRF optimization. Note that since \textit{sky} has no valid depth, we ignore the classified \textit{sky} pixels during occupancy mapping.

\subsection{Color and Label fusion}
\label{sec:color fusion}
Standard occupancy grid maps only store the occupancy value, however in our case, we also need to store the color and label distribution for latter CRF optimization. Since each grid can be observed by different pixels in different frames, we need to fuse the observation before optimization. For color fusion, we directly take the mean of different color observation in different views. For the label fusion, we follow the standard Bayes' update rule similar to that in occupancy mapping. Denote the label probability distribution of a grid at time $t$ as $x_t$ and the measurements till now as $z_{1:t}$. For the new combing image $z_t$, we first perform 2D semantic segmentation then update the 3D grid's label based on Bayes rule:
\begin{equation}
   \label{eq:prob fuse}
\begin{split}
p(x_t|z_{1:t}) &= \frac{p(x_t|z_t)p(z_t)}{p(x_t)} \frac{p(x_{t-1}|z_{1:t-1})}{p(z_t|z_{1:t-1})} \\
& \approx \frac{1}{Z} p(x_t|z_t)p(x_{t-1}|z_{1:t-1})
\end{split}
\end{equation}

\noindent To derive the second row, we can assume the class prior $p(x_t)$ as a constant and group $p(z_t)/p(z_t|z_{1:t-1})$ as the normalization constant. So for each incoming image, the label probability fusion is simply multiplication followed by normalization. The updated $p(x_t)$ will be further be optimized through CRF.

%% file: crf_model.tex

After building the 3D grid map, we utilize CRFs to jointly optimize each grid's label. We propose a general hierarchical CRF model with a robust $P^N$ potential and develop an efficient inference algorithm for it. Note that this CRF model can also be used for other problems such as 2D segmentation.

We start by defining each grid's label as a random variable $x_i$ taking a label from a finite label set $\mathcal{L}=\{l_1,l_2,...,l_k\}$ and the joint label configuration of all $N$ grids as $\mathbf{x} \in \mathcal{L}^N$. Then the joint probability and Gibbs energy is defined:
\begin{equation}
   \label{eq:prob energy}
    P(\mathbf{x}|\mathbf{D}) = \frac{1}{Z(\mathbf{D})} \exp(-E(\mathbf{x}|\mathbf{D}))
\end{equation}
\begin{equation}
   \label{eq:total energy}
   \begin{split}
    E(\mathbf{x|\mathbf{D}}) = \sum_i \psi_i^U(x_i)+  & \sum_{i<j}  \psi_{ij}^P(x_i,x_j) + \psi_{\mathbf{c}}^{HO}(\mathbf{x_c})
\end{split}
\end{equation}
where $P(\mathbf{x}|\mathbf{D})$ is the posterior probability of configuration $\mathbf{x}$ given a grid data $\mathbf{D}$, $Z(\mathbf{D})$ is the partial function for normalization. $\psi_i^U$ and $\psi_{ij}^P$ are the unary and pairwise potential energy. $\psi_\mathbf{c}^{HO}$ is the high order energy formed by all cliques $\mathbf{x_c}$. We will explain these three potentials in more details. Now the CRF inference problem of maximizing $P(\mathbf{x}|\mathbf{D})$ changes to find $\mathbf{x^*} = \arg\min_{\mathbf{x}} E(\mathbf{x}|\mathbf{D})$.

\subsection{Unary potential}
\label{sec:kitti crf separate crf unary}
Unary $\psi_i^U(x_i)$ represents the cost of a grid taking label $x_i $, which can also be treated as the prior distribution of variables. It is computed by the negative logarithm of the prior probability:
\begin{equation}
\label{eq:prob to unary}
\psi_i^U(x_i) = -\log p(x_i)
\end{equation}

\noindent where $p(x_i)$ is the fused label distribution from Equation \ref{eq:prob energy} in Section \ref{sec:color fusion}. There are many approaches on 2D semantic segmentation for example TextonBoost \cite{shotton2008semantic} and more recent CNN \cite{long2014fully}\cite{chao2016pop}. We adopt the dilated CNN \cite{yu2015multi} simply because it directly provides the trained model. Note that it doesn't contain post optimization such as CRF so the prediction might contain discontinuity and inconsistency between pixels as shown in Fig. \ref{fig:kitti crf whole system}(c).

\subsection{Pairwise potential}
\label{sec:separate crf pairwise}
The pairwise potential $\psi_{ij}^P(x_i, x_j)$ is adopted from \cite{krahenbuhl2012efficient}, defined as a combination of gaussian kernels:
\begin{equation}
   \label{eq:pairwise-potential}
    \psi_{ij}^P(x_i, x_j) = \mu (x_i, x_j) \sum_{m=1}^K w^{(m)}k^{(m)}(\textbf{f}_i, \textbf{f}_j)
\end{equation}

\noindent where $\mu (x_i, x_j)$ is the label compatibility function. We use the simple Potts model $\mu(x_i, x_j) = [x_i \neq x_j]$, which introduces a penalty for nodes assigned different labels. Each $k^{(m)}$ is a Gaussian kernel depending on feature vectors $\textbf{f}$. We currently use two kernel potentials defined based on the color and position of 3D grid:

\begin{equation}
   \label{eq:kernel}
	\begin{aligned}   
    k(\textbf{f}_i, \textbf{f}_j) &= w_1\exp\Big( -\frac{|p_i -p_j| }{2\theta_\alpha^2} - \frac{|I_i - I_j |}{2 \theta_\beta^2} \Big) \\
    &+w_2 \exp\Big(  -\frac{|p_i -p_j| }{2\theta_\gamma^2}  \Big) 
	\end{aligned}        
\end{equation}

There are also some other features in 3D space for example surface normal \cite{vineet2015incremental}. However, our occupancy grid is not dense and smooth enough to compute high quality surface normals.

\subsection{Hierarchical CRF Model}
\label{high_order}
The above dense pairwise potential lacks the ability to express more complicated and meaningful constraints. For example in a 2D image, pixels within a homogeneous region are likely to have the same label. Thus, we design high-order potentials $\psi_{c}^{HO}(x_c)$ to represent these constraints.

\begin{figure}[t]
  \centering     
  \subfigure[]
    {  \label{fig:robust pn}
       \includegraphics[scale=0.26]{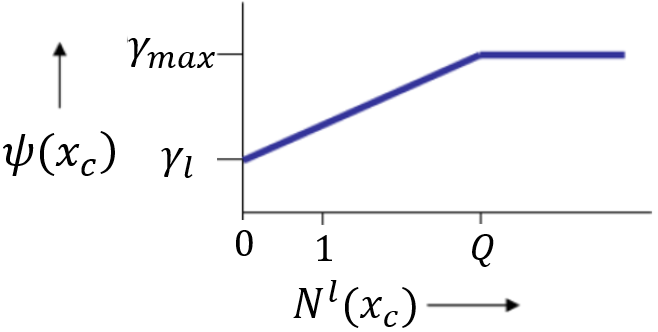}
    }  
  \subfigure[]
    {  \label{fig:auxiliary crf}
       \includegraphics[scale=0.19]{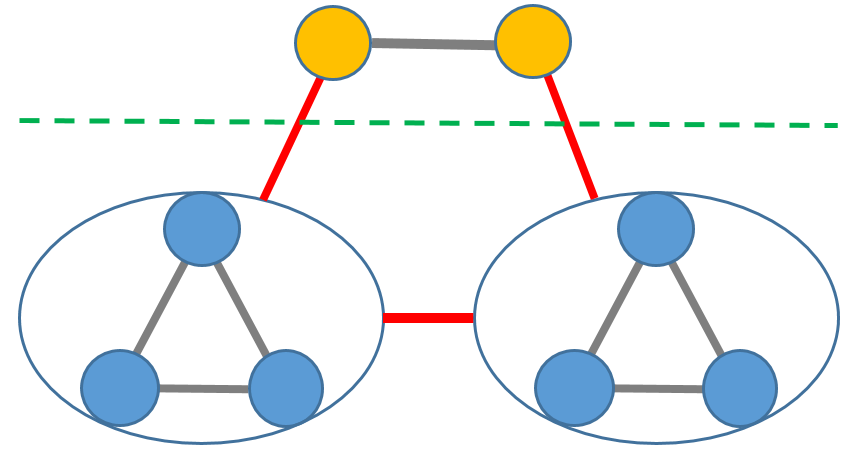}
    }
   \caption{(a) Robust $P^n$ Model cost for clique $c$ from \cite{kohli2007p3}. (b) Our hierarchical CRF model by representing clique as an yellow auxiliary node. The auxiliary clique node has dense pairwise connections between each other and also has connections to its children nodes. Best viewed in color.}
\end{figure}

There are different models for high order potentials. The $P^n$ Potts model \cite{kohli2007p3} rigidly enforces the nodes within a clique to take the same label which might be wrong due to inaccurate clique segmentation. Kohli \textit{et al.} \cite{kohli2009robust} then propose a Robust $P^n$ model whose cost is dependent on the number of variables taking the dominant label shown in Fig. \ref{fig:robust pn}. It could be treated as a soft version of $P^n$ Potts model. Robust $P^n$ is further shown to be equivalent to the minimization of a hierarchical pairwise graph with new added auxiliary variable $y_c$ representing the clique's label \cite{russell2009associative} shown as the yellow nodes in Fig. \ref{fig:auxiliary crf}. The robust $P^n$ potential for this clique is then defined as: 

\begin{equation}
\label{eq:ho potential}
\begin{aligned}
\psi^{HO}_c(x_c) &= \min_{y_c} \psi_c(x_c,y_c) \\
& = \min_{y_c}\big( \psi_c(y_c) + \sum_{i\in c} \psi_{ci}(y_c,x_i) \big)
\end{aligned}
\end{equation}

\noindent where $\psi_c(y_c)$ represents the auxiliary clique variable's unary cost. A separate classifier could be trained to compute it or we can simply use the mean of its children nodes' unary. $\psi_{ci}(y_c,x_i)$ encourages the consistency between clique variable and its children nodes shown as the edge between yellow nodes and blue nodes in Fig. \ref{fig:auxiliary crf}, defined as:

\begin{equation}
   \label{eq:crf clique pixel}
\psi_{ci}(y_c,x_i) = 
\begin{cases}
	0     & \quad \text{if } y_c=x_i\\
    k^l_c     	& \quad \text{Otherwise}\\
\end{cases}
\end{equation}

We further extend it to model the relationships between all auxiliary variables $y$ as a gaussian pairwise potential defined in Section \ref{sec:separate crf pairwise}, shown as the edges between yellow nodes. It basically encourages the label consistency between similar segment cliques. The clique variable's feature is computed by the mean of RGB and position of all its children nodes. 

In summary, let $V$ represents the set of low-level grid variables and $\mathcal{S}$ be the high-level clique variables. Then the total energy $E(\mathbf{x})$ in Equation \ref{eq:total energy} is transformed into:
\begin{equation}
\label{eq:hierarchical crf potential}
\begin{aligned}
E(\mathbf{x}) = & \sum_{i \in \mathcal{V}} \psi_i(x_i) + \sum_{i,j \in \mathcal{V}}\psi_{ij}(x_i,x_j) + \\
& \min_{\mathbf{y}}\Big(  \sum_{c\in \mathcal{S}} \psi_c(x_c,y_c) + \sum_{c,d \in \mathcal{S}} \psi_{cd}(y_c,y_d) \Big)
\end{aligned}
\end{equation}

\noindent where the first row defines the unary and pairwise potentials of low-level grid variables and the second row defines the high order potential of cliques and between cliques. Now this model is only composed of unary and pairwise terms but with more auxiliary variables.

\subsection{Mean Field Inference for Hierarchical CRFs}

\subsubsection{Algorithm Description}
Mean field inference for CRFs with $P^n$ potts model has been addressed by \cite{vineet2014filter}. In this section, we develop efficient inference for our hierarchical CRF model with Robust $P^n$ model.

We approximate the target distribution $P(\mathbf{x})$ using $Q(\mathbf{x})$ with the form $Q(\mathbf{x}) = \prod_i Q_i(x_i)$, namely all variables are marginally independent. During each iteration, we first update the distribution of high-level clique variables $y_c$ using Equation \ref{eq:kitti crf mean update} and find the MAP label assignment for them then update low-level grid nodes $x_i$ using Equation \ref{eq:kitti crf mean update2}. The two mean field update rules are as follows:

\begin{equation}
   \label{eq:kitti crf mean update}
   \begin{aligned}    
     Q^{t+1}(& y_c=l) = \frac{1}{Z_c} \exp \bigg( -\psi_c(y_c) - \sum_{d\in \mathcal{S},d \neq c}  \\
      & Q_d(y_d)\psi_{cd}(y_c,y_d) - \sum_{i\in c }Q_i(x_i)\psi_{ci}(y_c,x_i) \bigg)
     \end{aligned}
\end{equation}

\begin{equation}
   \label{eq:kitti crf mean update2}
   \begin{aligned}   
    Q^{t+1}(&x_i=l) = \frac{1}{Z_i} \exp \bigg( -\psi_i(x_i)  - \sum_{j\in \mathcal{V},j \neq i}\\
& Q_j(x_j)\psi_{ij}(x_i,x_j) - Q_c(y_c)\psi_{ci}(y_c,x_i) \bigg)
     \end{aligned}
\end{equation}

\subsubsection{Complexity Analysis}
\label{sec:kitti crf complexity analysis}
There are two main parts in the previous equation: dense pairwise terms $\psi_{cd}(y_c,y_d)$, $\psi_{ij}(x_i,x_j)$, and clique-children terms $\psi_{ci}(y_c,x_i)$.

For the dense pairwise parts, as shown by Kr{\"a}henb{\"u}hl \cite{krahenbuhl2012efficient}, using the technique of Gaussian convolutions and permutohedral lattice, the time complexity of the Potts model is $\mathcal{O}(KNL)$, where $K$ is the number of kernels, $N$ is total grid number and $L$ is the labels. So in our case the computation would be $\mathcal{O}(KN_gL+KN_cL)$, where $N_g=|V|$, $N_c=|S|$ are the number of low-level grids and high-level cliques respectively.

For the clique-children terms, we need to visit each low-level grid to check its label compatibility with the clique variable. In our setting, each grid only has at most one parent clique so the computation complexity would be $\mathcal{O}(N_g)$.

In all, the total time complexity is $\mathcal{O}(KN_gL+KN_cL+N_g) \approx \mathcal{O}(KN_gL)$ as clique number $N_c$ is usually much smaller than grid number $N_g$. Therefore, it has the same algorithm complexity with dense pairwise CRF in theory, which is linear in the number of grid $N_g$.



\if
The general form of mean field update is:
\begin{equation}
   \label{eq:mean_update update}
     Q^{t+1}(x_i=x)=
\frac{1}{Z_i} \exp \left( -\psi_i(x_i) -\sum_{c \in \mathcal{C}} \sum_{\mathbf{x}_c \mid x_i=x} Q^t(\mathbf{x}_{c-i}) \psi_c(\mathbf{x}_c) \right)
\end{equation}
\noindent Where $\mathcal{C}$ denotes all the cliques including pairwise terms and high order. $Q_{c-i}$ denotes the marginal distribution of all variables in $\mathcal{C}$ apart $x_i$.
\begin{equation}
   \label{eq:mean_update update}
     Q^{t+1}(x_i=l)=
\frac{1}{Z_i} \exp \left( -\psi_i(x_i) -\sum_{c \in \mathcal{C}} \sum_{\mathbf{x}_c \mid x_i=l} Q^t(\mathbf{x}_{c-i}) \psi_c(\mathbf{x}_c) \right)
\end{equation}

For high order terms, we follow a similar approach to \cite{vineet2014filter}, 
\begin{equation}
   \label{eq:high-order-update}
   \begin{split}   
	&\sum_{\{ \textbf{x}_c | x_i = l\}} Q_{c-i}(\textbf{x}_{c-i}) \dot \psi_c^{HO}(\textbf{x}_c) \\
	=& - \bigg ( \prod_{j \in c, j \neq i} Q_j (x_j = l) \bigg) \gamma_l s_{cl} \\
	&- \bigg ( 1 - \bigg(\prod_{j \in c, j \neq i} Q_j (x_j = l) \bigg) \bigg)\gamma_{min}
  \end{split}   
\end{equation}

\noindent which contribute $O(L \max_c (|c|) |C|)$ to each parallel update. Here, $|c|$ is the size of a clique and $|C|$ is the number of cliques. Thus, in our high order potential setting, we achieve the time complexity linear to the number of 3D grid $N$ which is the same as the dense pairwise updates.
\fi

%% file: result.tex
\subsection{Dataset and implementation}
We evaluate our system on two labelled datasets from KITTI \cite{geiger2012we}: Sengupta \cite{sengupta2013urban} containing 25 test images from sequence 15, and Kundu \cite{kundu2014joint} containing 40 test images mainly from sequence 05. They are also used in other state-of-the-art semantic 3D mapping system \cite{vineet2015incremental} \cite{sengupta2015semantic} \cite{sengupta2013urban}. In total, there are 11 object classes including building, road, car and so on.

As explained in Section \ref{sec:kitti crf separate crf unary}, our unary prediction comes from the dilation CNN \cite{yu2015multi} which is trained on another sequence of KITTI dataset. Fine-tuning on the experimental dataset could also be used to further improve the accuracy. We modify the scrolling occupancy grid library \cite{scrollgrid} to maintain a 3D map. The map size is $25\times25\times8m^3$ around the camera with grid resolution $0.1m^3$. Larger grid volume and finer grid resolution could improve the performance but the computation time also increases quickly. Grid cells outside of the bounding area will be removed to save memory and computation for online updates, but could also be stored in memory so as to create a final complete 3D map.

We use the SLIC algorithm to generate around 150 superpixels per image, shown in Fig. \ref{fig:kitti crf whole system}d. We transfer the pixel-superpixel membership to 3D space by back projecting pixels to the corresponding 3D grids.

To evaluate the segmentation accuracy, we project 3D labelled grid maps onto the camera image plane, ignoring grids that are too far from the camera. We choose $40$ meters as the threshold, instead of the $25$ meters in \cite{vineet2015incremental}. Though further grids have larger position uncertainty and reduce the segmentation accuracy, they can improve the mapping density and can also be beneficial for other applications.

\subsection{Qualitative Results}
\label{sec:kitti crf qualitative}

\begin{figure}
\vspace{1.5 mm}
    \centering
    \includegraphics[width=3.3in,height=1.25in]{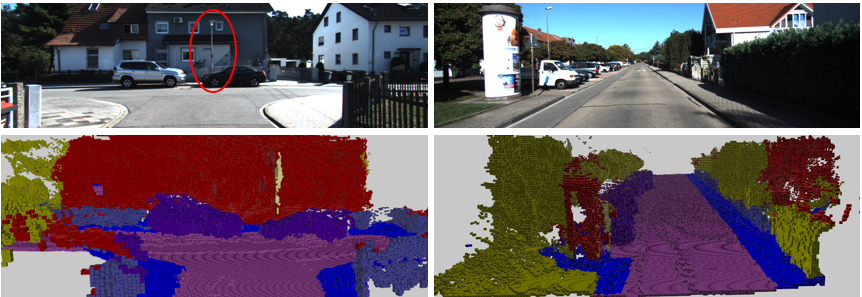}
    \caption{Qualitative results of our 3D semantic mapping. First row: input images. Second row:  semantic reconstruction scene. A thin pole is marked in red and also constructed in the map. In the second image, the building and car are clearly recognized.
}
    \label{fig:kitti crf qualitative result}
\end{figure}

\begin{figure*}
\vspace{1.5 mm}
  \centering
    \includegraphics[width=4.6in,height=1.40in]{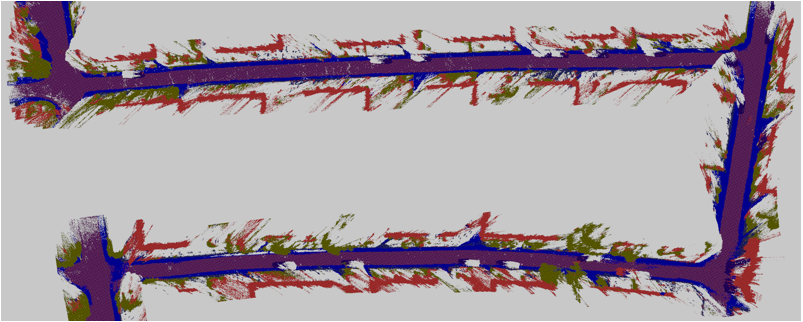}
    \includegraphics[width=0.99\textwidth,height=1.5in]{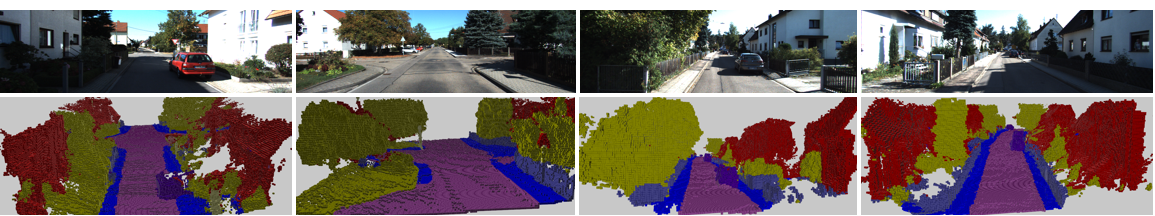}
   	\includegraphics[scale=0.45]{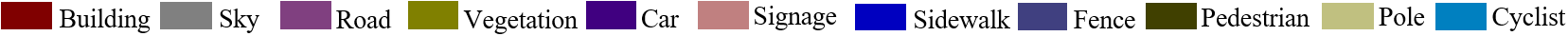}
  \caption{Visualization of 3D semantic mapping. (a) top view of 3D mapping over long sequence (850 images) of KITTI sequence 5. It demonstrates that our algorithm work well in large scale environments. (b) More 3D reconstruction examples with different scenarios.}
   \label{fig:more 3d example}  
\end{figure*}

We first present some qualitative results of semantic 3D mapping in Fig. \ref{fig:kitti crf qualitative result}. Our approach can successively recognize and reconstruct classes of general objects, and even thin objects such as poles in spite of the shadows and textureless surfaces. The top view of the whole sequence (keep all grids in memory) and more examples are shown in Fig. \ref{fig:more 3d example}.

\begin{figure}[h]
  \centering     
  \subfigure[]
    {  \label{fig:kitti raw image}
       \includegraphics[width=1.64in,height=0.7in]{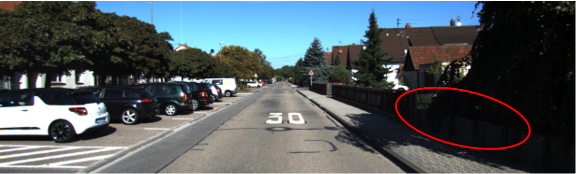}
    }\hspace{-3mm}       
  \subfigure[]
    {  \label{fig:kitti cnn 2d}
       \includegraphics[width=1.64in,height=0.7in]{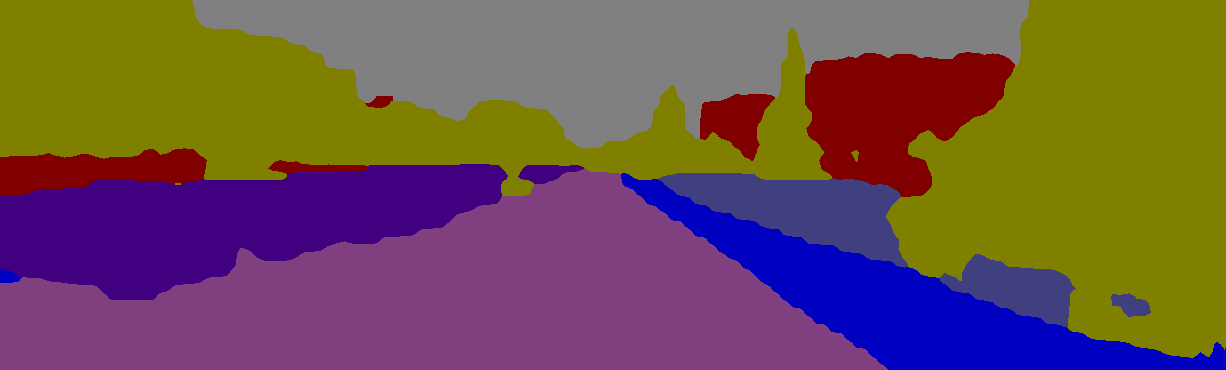}
    }\vspace{-2.5mm}
  \subfigure[]
    {  \label{fig:kitti 2d crf}
       \includegraphics[width=1.64in,height=0.7in]{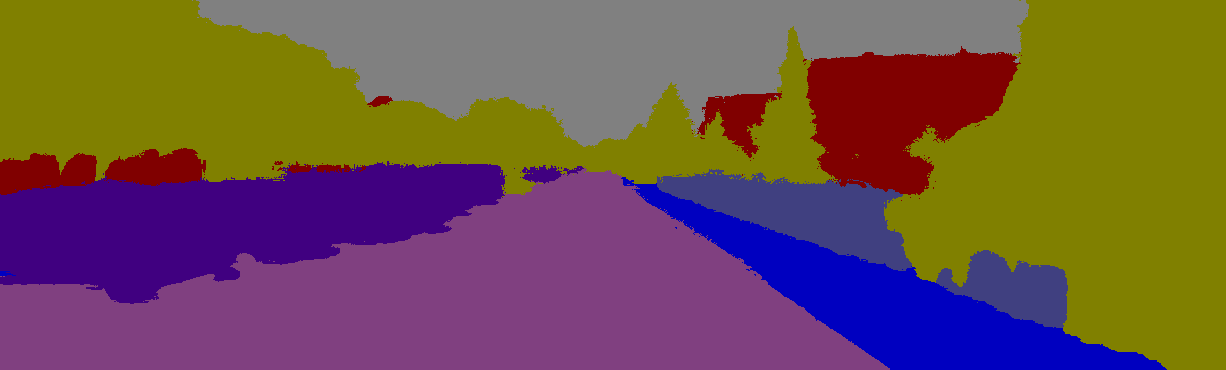}
    }\hspace{-3mm}
  \subfigure[]
    {  \label{fig:kitti 3d crf reproj result}
       \includegraphics[width=1.64in,height=0.7in]{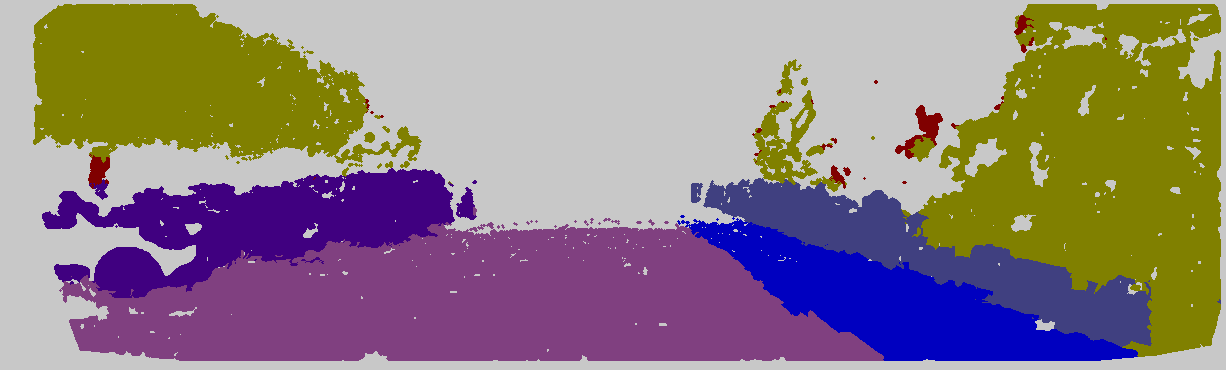}
    }
   \caption{Examples demonstrating the advantage of 3D optimization. (a) Raw RGB image (b) 2D CNN prediction  (c) 2D CRF optimization. \emph{fence} on the bottom right is still wrongly labelled due to the dark shadow of trees. (d) 3D CRF optimization result (back-projection onto image plane). Gray area means no projections. Fence is now correctly labelled.}
    \label{fig:kitti crf fence result}      
\end{figure}

In another example of Fig. \ref{fig:kitti crf fence result}, we demonstrate the advantage of 3D segmentation compared to 2D. In Fig. \ref{fig:kitti raw image}, due to the dark shadow of trees, \emph{fence} in the red eclipse area has similar intensity and texture with its surrounding trees thus it is very difficult to label them only from 2D image shown in Fig. \ref{fig:kitti cnn 2d} and Fig. \ref{fig:kitti 2d crf}. However, in 3D space, \emph{fence} is continuous and has different shape and position compared to trees and roads thus it can be correctly labelled using 3D optimization shown in Fig. \ref{fig:kitti 3d crf reproj result}.

\subsection{Quantitative Results}
In this section, we quantitatively evaluate and compare the segmentation accuracy with other approaches.

\begin{table*}
\vspace{1.5 mm}
\caption{Comparison of 2D/3D approaches on KITTI Sengupta (seq15) dataset}
\begin{center}
\begin{tabular}{l|ccccccccc|cc}
\hline
\multicolumn{3}{c|}{Method} 	&\rotatebox{90}{Building}  & \rotatebox{90}{Vegetation}	&\rotatebox{90}{Car}  &\rotatebox{90}{Road}		&\rotatebox{90}{Fence}  &\rotatebox{90}{Sidewalk}  &\rotatebox{90}{Pole}  &\rotatebox{90}{Average} &\rotatebox{90}{Global} 	 \\  \hline
\multirow{6}{*}{Accuracy} &\multirow{2}{*}{2D} 		&\multicolumn{1}{|c|}{Texton\cite{shotton2008semantic}}   
&97.0 &93.4 &93.9 &98.3 &48.5 &\textbf{91.3} &49.3 &81.7 &88.4  \\
						 &	  &\multicolumn{1}{|c|}{CNN} 
&\textbf{98.5} &\textbf{97.6} &\textbf{96.5} &\textbf{98.4} &\textbf{77.8} &87.9 &\textbf{51.1}&\textbf{86.8}&\textbf{94.2}\\\cline{2-12}
						 &\multirow{4}{*}{3D} 	&\multicolumn{1}{|c|}{Sengupta\cite{sengupta2013urban}}  
&96.1 &86.9 &88.5 &97.8 &46.1 &86.5 &38.2 &77.2 &85.0  \\
						 & 		&\multicolumn{1}{|c|}{Valentin\cite{valentin2013mesh}} 	
&96.4 &85.4 &76.8 &96.9 &42.7 &78.5 &39.3 &73.7 &84.5  \\
						 & 					&\multicolumn{1}{|c|}{Sengupta\cite{sengupta2015semantic}}	 
&89.1 &81.2 &72.5 &97.0 &45.7 &73.4 &3.30 &66.0 &78.3  \\
						 &  					&\multicolumn{1}{|c|}{Vineet\cite{vineet2015incremental}}	 
&97.2 &94.1 &94.1 &98.7 &47.8 &91.8 &51.4 &82.2 &---  \\
						 &  					&\multicolumn{1}{|c|}{Our before CRF}	 
&98.2 &98.5 &\textbf{96.3} &\textbf{99.5} &81.8 &90.2 &57.1 &88.9 &95.1   \\
						 &  					&\multicolumn{1}{|c|}{Our after CRF}	 
&\textbf{98.2} &\textbf{98.7} &95.5 &98.7 &\textbf{84.7} &\textbf{93.8} &\textbf{66.3} &\textbf{90.9} &\textbf{95.7} \\   \hline \hline
\multirow{6}{*}{IoU} & \multirow{2}{*}{2D} &\multicolumn{1}{|c|}{Texton\cite{shotton2008semantic}}   
&86.1 &82.8 &78.0 &\textbf{94.3} &47.5 &73.4 &39.5 &71.7 &\\
						 &					   &\multicolumn{1}{|c|}{CNN} 
&\textbf{93.3} &\textbf{89.8} &\textbf{94.1} &93.4 &\textbf{76.4} &\textbf{80.0} &\textbf{44.1} &\textbf{81.7} & \\  \cline{2-12}
						 & \multirow{4}{*}{3D} &\multicolumn{1}{|c|}{Sengupta\cite{sengupta2013urban}}  
&83.8 &74.3 &63.5 &96.3 &45.2 &68.4 &28.9 &65.7 & \\
						 & 					  &\multicolumn{1}{|c|}{Valentin\cite{valentin2013mesh}} 		
&82.1 &73.4 &67.2 &91.5 &40.6 &62.1 &25.9 &63.2 & \\
						 & 					&\multicolumn{1}{|c|}{Sengupta\cite{sengupta2015semantic}}	 
&73.8 &65.2 &55.8 &87.8 &43.7 &49.1 &1.9 &53.9 &\\
						 &  					&\multicolumn{1}{|c|}{Vineet\cite{vineet2015incremental}}	 
&88.3 &83.2 &79.5 &94.7 &46.3 &73.8 &41.7 &72.5 & \\
						 &  					&\multicolumn{1}{|c|}{Our before CRF}	 
&94.2 &90.8 &\textbf{95.4} &95.2 &79.9 &84.8 &54.2 &85.2 &  \\
						 &  					&\multicolumn{1}{|c|}{Our after CRF}	 
&\textbf{95.4} &\textbf{91.0} &94.6 &\textbf{96.6} &\textbf{81.1} &\textbf{90.0} &\textbf{61.5} &\textbf{87.6} & \\   \hline
\end{tabular}
\label{tab:kitti compare Sengupta result}
\end{center}
\vskip -0.1in
\end{table*}

\subsubsection{Comparison with 3D system}
As mentioned in Section \ref{sec:kitti crf review}, different systems take different 3D geometric mapping approaches such as voxel-hashing \cite{vineet2015incremental}, octomap \cite{sengupta2015semantic} and our occupancy grid, therefore a common approach for comparison is to project the 3D map onto image planes. We adopt the standard metric of (class) pixel accuracy defined as TP/(TP+FP) and (class) intersection over union (IoU) defined as TP/(TP+FP+FN). T/F P/N represents true/false positive/negative.

Comparison results are shown in Table \ref{tab:kitti compare Sengupta result}. \textit{Global} represents the overall pixel accuracy and \textit{Average} represents mean class accuracy or IoU. '---' indicates number not provided. In total, there are 11 class in the original dataset and we select 7 common class appearing in other systems for comparison. The results of other work are taken from the paper directly. Since \cite{sengupta2015semantic} and \cite{valentin2013mesh} also evaluate the \textit{signage} class which is ignored here, we recompute the metric \textit{Average} of their methods. Metric \textit{Global} cannot be re-computed as it depends on the actual distribution but it only has about $0.1\%$ variation because \textit{signage} occupies a very small area.

From the table, our method greatly outperforms other 3D systems in all categories. For the class \textit{fence}, there is a significant accuracy increase of 36.3\% over the state-of-the-art, and for the class \emph{pole}, we achieve a 10.3\% improvement. Overall, the mean class IoU is increased by 15\% and global pixel accuracy is increased by 10\%.

There are two main reasons why our approach outperforms existing systems. First, we use the latest the 2D CNN semantic segmentation as CRF unary shown as row 'CNN' in the table, while existing approaches all use TextonBoost \cite{shotton2008semantic} shown as row 'Texton'. CNN is much more accurate than TextonBoost nearly in all categories. Even using the simple label fusion in Section \ref{sec:color fusion} without further CRF optimization, the result in row \textit{Our before CRF} is already more accurate than other systems. Second, we propose a new hierarchical CRF model to optimize 3D grid labels which further improves the mean IoU by 2.4\% and global accuracy by 0.6\%. In some class such as \textit{Sidewalk} and \textit{Pole}, 3D CRF optimization has about 6\% improvement of IoU, which also matches the example in Fig. \ref{fig:kitti crf fence result}.

\subsubsection{Comparison of different CRF model}
In this section, we compare different CRF models to demonstrate the advantage of our proposed hierarchical model. As mentioned in Section \ref{sec:kitti crf semantic map}, our CRF model can also be applied to other optimization problems not limited to 3D mapping. So we evaluate the CRF on a 2D semantic segmentation task because it is not affected by other factors such as 3D reconstruction error, grid map resolution etc. We choose a more diverse Kundu dataset \cite{kundu2014joint} for evaluation.

\begin{figure*}
  \centering
  \includegraphics[width=6.6in,height=3.0in]{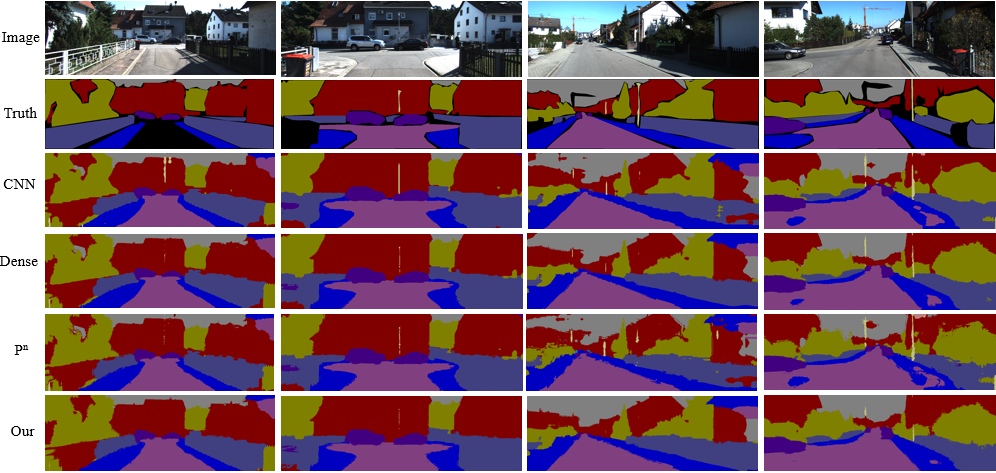}
  \caption{ Comparison of different CRF model on 2D semantic segmentation task in KITTI sequence 5. \textit{Dense} stands for dense CRF. $P^N$ stands for high order CRF with $P^N$ Potts model. The last line is our proposed hierarchical CRF with robust $P^N$ model. The black area in truth image is not labelled. Our method on the bottom row has better overall performance for example the sky region in the third image and road on the last image.}
  \label{fig:crf 2d example}    
\end{figure*}

We compare with two other popular CRF models: dense CRF \cite{krahenbuhl2012efficient} and $P^N$ Potts model \cite{vineet2014filter}, also adopted in other 3D mapping system. All the CRF models utilize the CNN prediction as unary cost then optimize for 5 iterations. Some qualitative comparison are shown in Fig. \ref{fig:crf 2d example} while quantitative result is shown Table \ref{tab:kitti kundu 2d result}. F.W. IoU stands for frequency weighed IoU. We can see that our model has higher IoU and global accuracy. For class \textit{sign}, we increase 10.5\% IoU compared to other models. This is mainly because that \textit{sign} is composed of triangular or square plates which usually form separate superpixels, thus suitable for our CRF model. However, some thin long objects such as \textit{Pole} usually break to different superpixels, making it difficult to optimize in 2D. However, in 3D space shown in previous Table \ref{tab:kitti compare Sengupta result}, our CRF model can still work well because even \textit{pole} is broken into multiple superpixels, its 3D position is quite different from its surroundings, therefore can still easily be classified.

\begin{table}[t]
\caption{Comparison of 2D CRF model on KITTI Kundu (seq05) dataset (IoU)}
\begin{center}
\begin{tabular}{l|ccccr}
\hline
Class			& CNN 		 	& Dense CRF		& $P^N$ Potts & Our hier					\\   \hline
Building  		& 83.8			& 85.4			& 84.5     	 & \textbf{86.6}					\\
Sky  			& 87.6			& 90.3			& 89.7     	 & \textbf{91.8}					\\
Road	    			& 90.0			& \textbf{90.2} 	& 90.0 		 & 90.1					\\
Vegetation		& 83.0			& 83.9			& 82.9    	 & \textbf{83.9}					\\
Sidewalk	  		& 74.4			& \textbf{74.7}	& 74.5	   	 & 74.3					\\
Car	  			& 72.5			& \textbf{73.8}	& 73.0	   	 & 73.1 					\\
Sign	  			& 23.1			& 29.3			& 24.7	   	 & \textbf{40.6}					\\  
Fence	  		& 69.5			& \textbf{70.4}	& 69.5	   	 & 70.1					\\  
Pole				& \textbf{33.6}	& 30.6			& 32.4		 & 23.5					\\  \hline
Mean IoU			& 68.6			& 69.6			& 69.0		 & \textbf{70.3}					\\
F.W IoU			& 81.5			& 82.6			& 81.9		 & \textbf{82.7}          \\
Global Acc		& 89.9 			& 90.5			& 90.1    	 & \textbf{90.6}					\\  \hline
\end{tabular}
\label{tab:kitti kundu 2d result}
\end{center}
\end{table}

\subsection{Time analysis}
We also provide time analysis of different components in our system. Except for the CNN unary prediction, all other computation are implemented on desktop CPU Intel i7-4790. CNN prediction time is not included here as it depends on the model complexity and GPU power. In fact, there are also some fast and lightweight CNNs that could even run in real time on embedded system \cite{paszke2016enet}.

There are five main components of our system shown in Table \ref{Table:kitti crf result time}. The first three parts utilize the existing public libraries and can further be computed in parallel threads to improve speed. From the label fusion in Section \ref{sec:color fusion}, the grid probability distribution from previous frames is  updated and further optimized by the current frame's observation. Since each frame only updates a small part of the 3D grid map, we only need to optimize for a few iterations until convergence. Through our test, even one single mean-field update is enough to produce good results. In Table \ref{tab:kitti compare Sengupta result}, the result is also reported where CRF iteration is set to one.

From the table, in the current settings, the grid optimization can run at 2Hz for hierarchical CRF or 2.5Hz without high order CRF potential, slower compared to the image frame rates (10Hz). As analyzed in  Section \ref{sec:kitti crf complexity analysis}, the computation $\mathcal{O}(KNL)$ is linear in the number of grids and label. There are several possible ways to improve the speed. Firstly, we can reduce the grid size or grid resolution to reduce the number of variables $N$ in the CRF optimization. However, this might increase the 3D reconstruction error and may not qualify for other needs. Secondly, reduce the number of labels $L$. If we only optimize for 7 main classes instead of current 11 classes, the speed can be almost doubled. Thirdly, GPU could also be used for parallel mean field updates which has been used in \cite{vineet2015incremental}. Lastly, we actually don't need to process each frame as adjacent frames may appear quite similar and have small affect the 3D map. Through our test, by processing every four frames to satisfy real-time rates, the global accuracy only reduces from $95.7$ to $95.3$ and IoU reduces from $87.6$ to $86.3$.

\begin{table}[t]
\vspace{1.5 mm}
\caption{Time analysis of our algorithm.}
\begin{center}
\begin{tabular}{lcccr}
\hline
Method 		    			& Time(s)  \\  \hline
Dense Disparity  		& 0.416    \\
State Estimation    		& 0.102    \\
Superpixel Generation  	& 0.115    \\
Occupancy Mapping     	& 0.295    \\
CRF w/o Hierarchical   	& 0.53/0.38	   \\  \hline
\end{tabular}
\label{Table:kitti crf result time}
\end{center}
\end{table}

%% file: conclusion.tex
In this paper, we presented a 3D online semantic mapping system using stereo cameras in (near) real time. Scrolling occupancy grid maps are used to represent the world and are able to adapt to large-scale environments with bounded computation and memory. We first utilize the latest 2D CNN segmentation as prior prediction then further optimize grid labels through a hierarchical CRF model. Superpixels are utilized to enforce smoothness. Efficient mean field inference for the high order CRF with robust $P^N$ potential is achieved by transforming it into a hierarchical pairwise CRF. Experiments on the KITTI dataset demonstrate that our approach outperforms the state-of-the-art approaches in terms of segmentation accuracy by 10\%.

Our algorithm can be used by many applications such as virtual interactions and robot navigation. In the future, we are interested in creating more abstract and high level representation of the environments such as planes and objects. Faster implementation on GPU could also be explored.